\documentclass[conference,10pt]{IEEEtran}
\IEEEoverridecommandlockouts
\usepackage{amsmath}
\usepackage{iftex}
\usepackage[T1]{fontenc}
\ifPDFTeX
  \usepackage[utf8]{inputenc}
\fi
\usepackage{newtxtext,newtxmath}
\usepackage[varqu,scaled=0.92]{zi4}
\ifPDFTeX\else
\fi
\usepackage[american]{babel}
\usepackage{booktabs,array,tabularx}
\usepackage{graphicx}
\usepackage{newfloat}
\DeclareFloatingEnvironment[name=Listing,placement=!t]{listing}
\usepackage{microtype}
\usepackage{xcolor}
\usepackage{cite}
\usepackage[hidelinks]{hyperref}
\usepackage{xurl}
\hypersetup{pdftitle={Explaining Process, Instrument and Data-Path Causes in Plain Language: Local Language Models That Use Detector Decisions},pdfauthor={Oleksandr Klymenko}}
\usepackage{placeins}
\usepackage{dblfloatfix}
\begin{document}

\title{Explaining Process, Instrument and Data-Path Causes in Plain Language: Local Language Models That Use Detector Decisions}

\author{\IEEEauthorblockN{Oleksandr Klymenko}
\IEEEauthorblockA{Independent Researcher, Poland \\
ORCID 0009-0000-5924-3697 \\
alexanderklmn@gmail.com}}

\maketitle

\begin{abstract}
Language models can turn the terse output of telemetry detectors into explanations an operator reads in passing, but they can also name the wrong cause, quote a number that was never measured, or recommend an action that does not fit. We separate the roles: deterministic detectors decide whether a departing reading reflects a process change, an instrument fault or data-path corruption; a local language model only explains the decision; and six mechanical grounding rules --- the stated cause must be the decided one, every number must occur in the evidence, the recommended check must belong to the cause --- admit or reject each explanation, with a deterministic template as the fallback. The explainer reaches the model through Spring AI or LangChain4j beneath a reliability layer. On 260 incidents from a published triage benchmark, four local models (3~B--32~B parameters, Ollama on one GPU) produced explanations that passed the rules in 1,296 of 1,300 cases; the share that quoted a number from the evidence rose from 14~\% for a 3~B model to 65~\% for a 32~B model, at 0.23~s and 15~s median latency respectively. In the author's rating of 50 accepted explanations, 46 were faithful to the evidence; all four that were not came from one 7~B model, through both frameworks. Asked instead to attribute the cause themselves from the same evidence, the models were correct in 47--59~\% of incidents, and from the raw readings in 5--41~\%, against 90.0~\% for the detectors; given readings alone, two models answered \emph{instrument} for nearly every incident. When the model server failed, every explanation fell back to the template within 10.3~s. The measurements support keeping the decision in the detectors and giving the model the wording.
\end{abstract}


\section{Introduction}

When a reading departs from expectation, an operator needs two things: a verdict --- did the process change, is the instrument failing, or was the value corrupted on its way? --- and a reason they can check. Deterministic detectors give the first and only fragments of the second: a CUSUM alarm, a GESD outlier count, a level that moved 7.43 $\sigma$ within five samples. A language model can turn such fragments into a sentence an operator reads in passing. It can also turn them into a confident sentence that names the wrong cause, a number that was never measured, or an action that does not fit.

Monitoring systems of the kind described in~\cite{r1,r2,r3} feed such channels to operators. This paper separates the two roles. The detectors of~\cite{r4}, whose components are also published in an open algorithm library~\cite{r5}, \emph{decide} --- they attribute each confirmed departure to process change, instrument fault, data-path corruption or \emph{undetermined}, on one channel, in constant memory. A language model running locally \emph{explains} --- it writes two or three sentences from the decision and its evidence. Between the model and the operator stands a set of grounding rules that every explanation must pass: the cause it states must be the decided cause, every number in it must occur in the evidence, the action it recommends must belong to that cause, and it must not assert another cause. An explanation that fails any rule is replaced by a deterministic template, as is any explanation whose model call fails. The model can therefore make the operator's text better; it cannot make the decision different, and it cannot make the system stop.

Two questions follow, and the paper answers both with measurements. \emph{Is the arrangement worth it} --- would the model not decide as well as the detectors, if asked? We give four local models the same evidence the detectors had, without their verdict, and alternatively the raw readings, and compare their attributions with the injected truth. \emph{Does the gate hold} --- how many explanations pass, which rules reject the others, and does the operator still get an explanation when the model server fails?

We contribute:

\begin{enumerate}
\item An \textbf{explain-only architecture} for model-assisted triage (Section III): a port with adapters for Spring AI and LangChain4j, six grounding rules, a template fallback, and observations of every outcome --- with the model reached through the reliability layer of~\cite{r6}.
\item A \textbf{measurement of the division of labour} (Section IV.C): attribution accuracy of four local models (3~B--32~B parameters) from evidence and from readings, against the detectors on the same 260 incidents.
\item A \textbf{measurement of the gate} (Section IV.B): acceptance, rejection by rule, specificity and latency of explanations from four models and two frameworks, with the author's rating of 50 accepted explanations; and the explainer's behaviour under provider failure (Section IV.D).
\item Open code and data~\cite{r7}.
\end{enumerate}

\section{Related work}

\textbf{Language models for incidents and anomalies.} Large language models have been used to recommend root causes and mitigations for cloud incidents~\cite{r8,r9}, and to detect anomalies in, or forecast, time series directly~\cite{r10,r11}. In these uses the model reaches the verdict. Results on time series are mixed: in~\cite{r10}, models detect some anomaly types but are outperformed by classical detectors. Our setting differs in that the verdict already exists, is produced by a method with a stated error rate~\cite{r4}, and the model's task is reduced to wording it.

\textbf{Grounding and hallucination.} Generated text that is fluent but unsupported by the input is the central failure of neural text generation~\cite{r12}; detecting it after the fact by sampling consistency~\cite{r13} or by checking atomic facts~\cite{r14} requires further model calls. Data-to-text generation has long faced the same problem with templates as the safe baseline~\cite{r15,r16}. We use the narrowness of the task to check grounding without a second model: the input is small and structured, the output is required in a fixed format, and the rules are mechanical.

\textbf{The data path.} Data-path corruption, the third cause, is not only a transmission error. The client stacks of industrial protocols convert and scale values between the wire and the application, and a defect there changes a reading while the instrument is sound --- for example, unsigned 64-bit values converted to floating point or subtracted incorrectly, which the author has corrected in the Eclipse Milo OPC UA stack (eclipse-milo/milo\#1960, shipped in Milo 1.1.7; eclipse-milo/milo\#1977, eclipse-milo/milo\#2029)~\cite{r17}. The acquisition path can also misbehave under load: the author has fixed a request throttle in Apache PLC4X that silently ignored a lowered limit while requests were in flight (apache/plc4x\#2734)~\cite{r18}. These are the situations in which the detectors' data-path verdict, and the operator's understanding of it, matter most.

\textbf{Placement of the model call.} The call to a model is itself a point of failure; the reliability layer of~\cite{r6} makes provider failures visible and bounded beneath Spring AI and LangChain4j. The explainer relies on it so that a failing model server degrades the explanation, not the triage.

\section{Method}

\begin{figure*}[!t]
\centering
\includegraphics[width=0.86\textwidth]{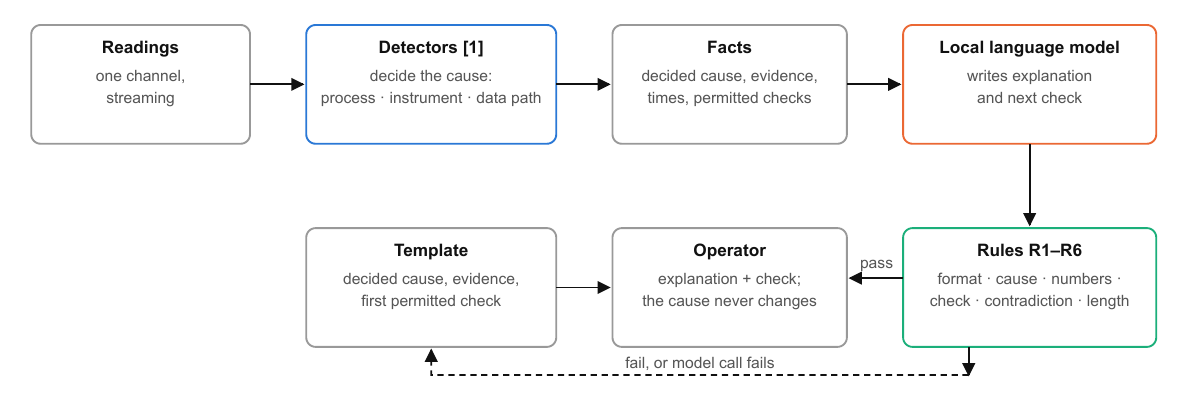}
\caption{Division of roles. The detectors of~\cite{r4} decide the cause; the model receives the decision and its evidence and writes the explanation; rules R1--R6 admit it or replace it with a template, which is also used when the model call fails. The cause shown to the operator is always the detectors'.}
\label{fig:1}
\end{figure*}

\subsection{Incidents}

An \emph{incident} is one decision of the triage method~\cite{r4} together with what the operator would see: the channel (a regulated tank level or a heated vessel's temperature), its declared range, its calibrated noise $\hat{\sigma}$, the decided cause, the detectors that fired, the evidence lines the rule produced, and the trigger and decision times. For evaluation, each incident also carries the fault that was injected and its true cause; neither is ever shown to a model.

Incidents are taken from the synthetic test runs of~\cite{r4} (seeds 2000--2049, both process models, the level channel alone, the frozen configuration): every detected event is paired with the first decision whose trigger falls in its window, the rule of the benchmark. The benchmark is used at the commit of the published version of~\cite{r4} (\texttt{3246e44}). Of 2,834 detected events, 20 per fault type were drawn at random (seed 20261006), 260 in all; 26 of them (10~\%) carry a decision that differs from the true cause. Development used 65 incidents (5 per type) from the method's development seeds 1000--1019, exported from the preceding commit of the benchmark (\texttt{6c3c65d}); the rules and prompts were frozen on them before the test incidents were drawn, and the changes made during development are listed in the repository (\texttt{docs/DEVELOPMENT.md}).

\subsection{Explanation and the grounding rules}

The model receives a system message stating its task and rules, and a user message with the incident's FACTS and a list of eight permitted next checks, two per cause, each with an identifier. It must reply with one JSON object: \texttt{cause}, \texttt{explanation}, \texttt{next\_check}. Decoding is greedy (temperature 0, fixed seed), at most 300 tokens.

A reply is shown to the operator only if it passes six rules:

\begin{itemize}
\item \textbf{R1 format} --- one JSON object with the three string fields.
\item \textbf{R2 cause} --- \texttt{cause} names the decided cause.
\item \textbf{R3 numbers} --- every number in the explanation equals a number in FACTS, or is that number rounded to the digits written: a token $x$ with $d$ decimals is grounded if
\begin{equation}
\exists\,a\in N(\text{FACTS}):\;|x-a|\;\le\;\tfrac12\,10^{-d}.
\label{eq:1}
\end{equation}
\item \textbf{R4 next check} --- \texttt{next\_check} is a permitted check \emph{for the decided cause}.
\item \textbf{R5 no contradiction} --- no sentence asserts another cause, by a phrase list per cause, unless the sentence negates it. For an \emph{undetermined} decision, naming the process and instrument hypotheses it cannot separate is allowed.
\item \textbf{R6 length} --- at most 80 words.
\end{itemize}

A reply that fails any rule, and a call that fails, yield the template: the decided cause, its meaning, the first two evidence lines, the times, and the first permitted check. The template passes all six rules by construction (a unit test checks this for every cause). Formally, let $F$ be an incident's facts, $r$ the model's reply, $R_i(F,r)\in\{0,1\}$ the six rules and $T(F)$ the template. The gate and the text the operator receives are

\begin{equation}
\begin{gathered} G(F,r)\;=\;\prod_{i=1}^{6} R_i(F,r),\\ E(F,r)\;=\;\begin{cases} r, & G(F,r)=1,\\ T(F), & \text{otherwise,}\end{cases} \end{gathered}
\label{eq:2}
\end{equation}

\noindent with a failed model call treated as $G=0$. Since $R_i(F,T(F))=1$ for every $i$, $E(F,r)$ satisfies all six rules whatever $r$ is: the properties hold by construction, not by the model's compliance. The rules guarantee properties an operator relies on --- the cause is the detectors', the numbers are the measurements, the action fits the cause --- but not that the prose is faithful in every respect: R3 checks that a number occurs in FACTS, not that it is used in its role. We measure what the rules let through separately (Section IV.B) and leave semantic faithfulness to a human audit.

\subsection{Implementation}

The explainer depends on a one-method port, \texttt{chat(system, user)}. Two adapters implement it: Spring AI's \texttt{OpenAiChatModel} and LangChain4j's \texttt{OpenAiChatModel}, both pointed at the OpenAI-compatible endpoint of a local Ollama server~\cite{r19}, both with their own retries off and the guard of~\cite{r6} installed (policy: three attempts, 120~s per response, 300~s in all). Each explanation is one Micrometer observation \texttt{triage.explain} with \texttt{source} (model or template), \texttt{rule} (the first failed rule) and \texttt{port}.

\subsection{Experiments}

\textbf{X1 --- explanations.} All 260 incidents, through Spring AI with \texttt{llama3.2} (3~B), \texttt{qwen2.5:7b}, \texttt{mistral:7b} and \texttt{qwen2.5:32b}, and through LangChain4j with \texttt{qwen2.5:7b}, on one RTX 3090 (Ollama 0.40.5; model digests \texttt{a80c4f17acd5}, \texttt{845dbda0ea48}, \texttt{6577803aa9a0} and \texttt{9f13ba1299af} respectively; runs of 18 October 2026). \textbf{X2 --- the model as attributor.} The same incidents and the four models through Spring AI, asked to attribute the departure themselves --- once from FACTS without the decided cause (the detectors' evidence), once from the 21--80 readings around the event alone --- with a definition of the four causes and a JSON reply. \textbf{X3 --- provider failure.} Both adapters against the scripted provider of~\cite{r6} answering 503 on every attempt, 429 with \texttt{Retry-After: 60}, nothing for 70~s, or a 200 that is not JSON; five incidents each, with a short policy (3~s per response, 60~s in all).

\subsection{Measures}

Let $\mathcal I$ be the $N=260$ test incidents, $c^\ast(i)$ the injected cause of incident $i$, $\hat c_D(i)$ the detectors' decision and $\hat c_m(i)$ the cause a model $m$ returns in X2. For X1, the \emph{acceptance rate} and the \emph{specificity} of model and framework $m$ are

\begin{equation}
\begin{gathered} a_m=\frac1N\sum_{i\in\mathcal I}G\big(F_i,r_m(i)\big),\\ s_m=\frac{\sum_{i} G\big(F_i,r_m(i)\big)\,\kappa\big(F_i,r_m(i)\big)}{\sum_{i} G\big(F_i,r_m(i)\big)}, \end{gathered}
\label{eq:3}
\end{equation}

\noindent where $\kappa=1$ if the explanation contains a number grounded by (1) in the \emph{evidence lines} of $F_i$, as opposed to the times, the range or the permitted checks. For X2, the \emph{attribution accuracy} and the \emph{agreement with the detectors} are

\begin{equation}
\begin{gathered} A_m=\frac1N\sum_{i\in\mathcal I}\mathbf{1}\big[\hat c_m(i)=c^\ast(i)\big],\\ Q_m=\frac1N\sum_{i\in\mathcal I}\mathbf{1}\big[\hat c_m(i)=\hat c_D(i)\big], \end{gathered}
\label{eq:4}
\end{equation}

\noindent and $A_D$ is the detectors' accuracy on the same incidents. Intervals are 95~\% percentile-bootstrap intervals over incidents (over accepted explanations for $s_m$), from $B=2000$ resamples with a fixed seed: $\big[\hat\theta^{\ast}_{(0.025B)},\,\hat\theta^{\ast}_{(0.975B)}\big]$, where $\hat\theta^{\ast}_{(k)}$ is the $k$-th smallest resampled value.

\section{Results}

\begin{table*}[!t]
\centering
\footnotesize
\setlength{\tabcolsep}{4pt}
\caption{Explanations of 260 incidents (X1).}
\label{tab:1}
\begin{tabularx}{\textwidth}{>{\raggedright\arraybackslash}p{3.1cm}>{\centering\arraybackslash}X>{\centering\arraybackslash}X>{\centering\arraybackslash}X>{\centering\arraybackslash}X>{\centering\arraybackslash}X}
\toprule
\textbf{Model \textperiodcentered{} framework} & \textbf{Accepted} & \textbf{Rejected (rule)} & \textbf{Quotes an evidence number (of accepted)} & \textbf{Words, median} & \textbf{Latency median / P95 (s)} \\
\midrule
llama3.2 (3~B) \textperiodcentered{} Spring AI & 260 & 0 & 36 (13.8~\%) & 11 & 0.23 / 0.31 \\
qwen2.5:7b \textperiodcentered{} Spring AI & 260 & 0 & 39 (15.0~\%) & 14 & 0.47 / 0.55 \\
qwen2.5:7b \textperiodcentered{} LangChain4j & 260 & 0 & 40 (15.4~\%) & 14 & 0.46 / 0.55 \\
mistral:7b \textperiodcentered{} Spring AI & 256 & 4 (R1) & 115 (44.9~\%) & 27 & 0.61 / 0.77 \\
qwen2.5:32b \textperiodcentered{} Spring AI & 260 & 0 & 168 (64.6~\%) & 30 & 14.7 / 18.2 \\
\bottomrule
\end{tabularx}
\end{table*}

\subsection{Overview}

All 1,300 explanation calls (X1), 2,080 attribution calls (X2) and 40 failure calls (X3) completed; no call to the working model server failed.

\begin{figure*}[!t]
\centering
\includegraphics[width=0.86\textwidth]{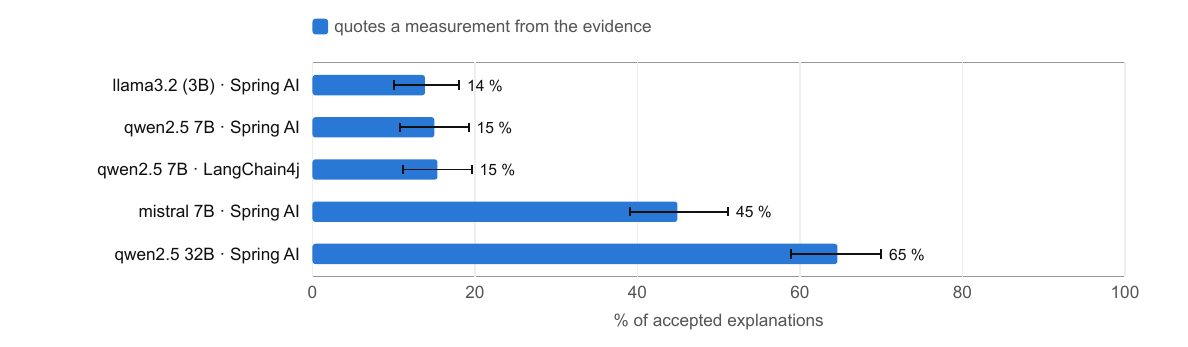}
\caption{Share of accepted explanations that quote a measurement from the evidence, per model and framework (X1, 260 incidents); whiskers are 95~\% percentile-bootstrap intervals.}
\label{fig:2}
\end{figure*}

\subsection{Explanations pass the rules; they differ in what they say}

With the rules stated in the prompt, the models almost never broke them: 1,296 of 1,300 explanations were accepted (Table~\ref{tab:1}). All four rejections are format failures by mistral:7b, which omitted the closing brace of the JSON object, each time on an \emph{undetermined} decision. No reply named a cause other than the decided one, quoted a number absent from the facts, or recommended a check for another cause. This holds also for the 26 incidents in which the decision was wrong: every accepted explanation stated the decided cause, as R2 requires, rather than correcting it.

What the rules do not constrain is content, and there the models differ by a factor of four. The two smaller models mostly restate the meaning of the cause --- ``The measured quantity itself changed, indicating a problem with the process'' (llama3.2) --- and quote a measurement in 14--15~\% of explanations; mistral:7b quotes one in 45~\%, and qwen2.5:32b in 65~\%, for example ``The vessel temperature changed by 4.23$\sigma$ from the baseline within a plausible rate, indicating a genuine process change'' (qwen2.5:32b). The larger model's specificity costs a 31-fold longer response than the 7~B model's on the same GPU.

The two frameworks are interchangeable behind the port: with the same model and greedy decoding, Spring AI and LangChain4j produced the same explanation text for 206 of 260 incidents, and the same acceptance and specificity within one incident.

\textbf{What the rules let through.} The rules check form, not meaning. To see what they admit, the author rated a random sample of 50 accepted explanations, 10 per model and framework, on two questions: is the explanation faithful to the decision and its evidence, and does it contain a claim the evidence does not support. The sheet with the ratings and the rater's notes is released with the code~\cite{r7}. 46 of 50 explanations were rated faithful, and 5 contained an unsupported claim. All four unfaithful explanations came from qwen2.5:7b, two through each framework, so the error belongs to the model, not to the framework: three described evidence of noise growth as a bias, one adding ``drift or stuck value'', which a stuck value would contradict, and the fourth described an abrupt step as ``bias or drift''. The fifth unsupported claim, from qwen2.5:32b on the same incident, listed every instrument mechanism --- bias, drift, noise growth, stuck value, saturation --- for that single step. mistral:7b and llama3.2 made no unsupported claim, but eight of llama3.2's ten explanations restated the definition of the cause without using the evidence: the generic text counted in Table~\ref{tab:1}. Three further patterns were noted without being rated unfaithful. Seven explanations of data-path decisions named one kind of corruption where the evidence lists three alternatives (``outside the declared range, missing, or at a boundary value''), which points to the wording of the evidence as much as to the models; one explanation of an \emph{undetermined} decision called it a sensor issue; and two qwen2.5:7b explanations quoted the 40~s between the first trigger and the decision as the duration of a stuck value.

\subsection{The models do not attribute as well as the detectors}

\begin{table}[!t]
\centering
\footnotesize
\setlength{\tabcolsep}{4pt}
\caption{Attribution by the models themselves (X2), correct / 260, against the injected truth; the detectors' decisions on the same incidents are correct in 234 (90.0~\%).}
\label{tab:2}
\begin{tabularx}{\columnwidth}{l>{\centering\arraybackslash}X>{\centering\arraybackslash}X>{\centering\arraybackslash}X}
\toprule
\textbf{Model} & \textbf{From the evidence} & \textbf{From the readings} & \textbf{Agrees with the detectors (evidence)} \\
\midrule
llama3.2 (3~B) & 134 (51.5~\%) & 98 (37.7~\%) & 158 (60.8~\%) \\
qwen2.5:7b & 154 (59.2~\%) & 99 (38.1~\%) & 162 (62.3~\%) \\
mistral:7b & 121 (46.5~\%) & 14 (5.4~\%) & 130 (50.0~\%) \\
qwen2.5:32b & 136 (52.3~\%) & 107 (41.2~\%) & 160 (61.5~\%) \\
\bottomrule
\end{tabularx}
\end{table}

\begin{table}[!t]
\centering
\footnotesize
\setlength{\tabcolsep}{4pt}
\caption{Correct attributions per true cause (process 60, instrument 100, data path 100 incidents).}
\label{tab:3}
\begin{tabularx}{\columnwidth}{l>{\raggedright\arraybackslash}X>{\centering\arraybackslash}X>{\centering\arraybackslash}X>{\centering\arraybackslash}X}
\toprule
\textbf{Attributor} & \textbf{Input} & \textbf{Process} & \textbf{Instrument} & \textbf{Data path} \\
\midrule
detectors~\cite{r4} & --- & 54 & 80 & 100 \\
llama3.2 & evidence & 54 & 60 & 20 \\
qwen2.5:7b & evidence & 38 & 87 & 29 \\
mistral:7b & evidence & 1 & 63 & 57 \\
qwen2.5:32b & evidence & 55 & 61 & 20 \\
llama3.2 & readings & 0 & 98 & 0 \\
qwen2.5:7b & readings & 0 & 99 & 0 \\
mistral:7b & readings & 12 & 0 & 2 \\
qwen2.5:32b & readings & 55 & 43 & 9 \\
\bottomrule
\end{tabularx}
\end{table}

\begin{figure*}[!t]
\centering
\includegraphics[width=0.86\textwidth]{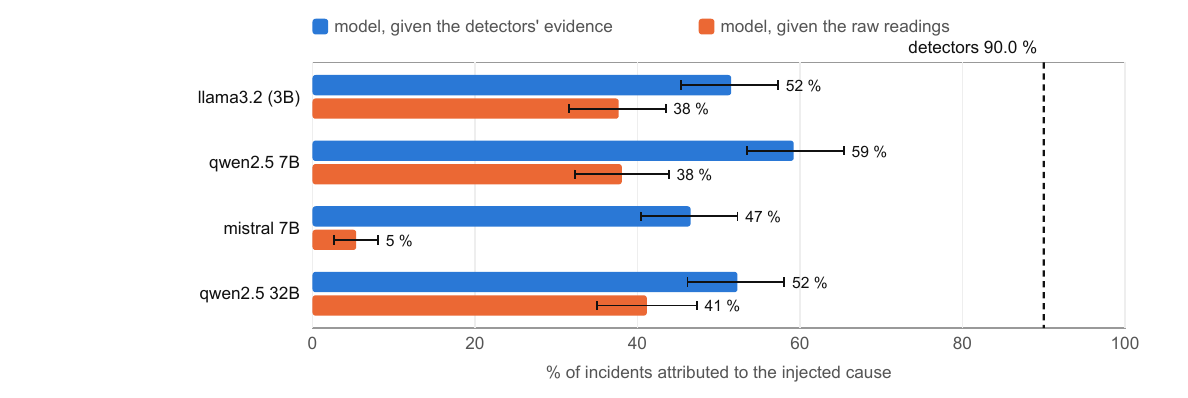}
\caption{Attribution by the models themselves (X2): from the detectors' evidence without their verdict, and from the raw readings, against the detectors' decisions on the same 260 incidents (dashed). Whiskers are 95~\% percentile-bootstrap intervals.}
\label{fig:3}
\end{figure*}

Given the detectors' evidence without their verdict, the best model attributed 59.2~\% of incidents correctly (Tables~\ref{tab:2} and~\ref{tab:3}, Fig.~\ref{fig:3}); the 32~B model did no better than the 3~B model (52.3~\% and 51.5~\%). The detectors were correct in 90.0~\%. The gap is largest for data-path corruption, which the detectors attribute correctly in all 100 incidents and the models in 20--57: an evidence line such as ``values outside the declared range, missing, or at a boundary value: 5'' names the data-path signature, but the models more often read it as an instrument failing.

Given only the readings, the models fell further, and the failure was systematic rather than noisy: llama3.2 answered \emph{instrument} for 258 of 260 incidents and qwen2.5:7b for 259, so that their 98--99~\% on instrument faults is the base rate of a constant answer; mistral:7b answered \emph{undetermined} for 190. Only the 32~B model spread its answers over the causes, for 41.2~\% correct.

Could the model at least repair the detectors' mistakes? On the 26 incidents with a wrong decision, the models attributed the true cause from the evidence in 1 (llama3.2, mistral:7b, qwen2.5:32b) and 9 (qwen2.5:7b) cases --- while, on the 234 incidents the detectors got right, qwen2.5:7b disagreed with them in 89. Replacing a decision by the model's whenever they differ would lose far more than it gains.

\subsection{When the model server fails, the operator still gets an explanation}

\begin{table}[!t]
\centering
\footnotesize
\setlength{\tabcolsep}{4pt}
\caption{Explanations under provider failure (X3; five incidents per cell; guard policy 3 attempts, 3~s per response, 60~s budget, \texttt{Retry-After} honoured up to 30~s).}
\label{tab:4}
\begin{tabularx}{\columnwidth}{l>{\centering\arraybackslash}X>{\centering\arraybackslash}X}
\toprule
\textbf{Failure} & \textbf{Spring AI: template / max time / attempts} & \textbf{LangChain4j: template / max time / attempts} \\
\midrule
503 on every attempt & 5/5 \textperiodcentered{} 2.1~s \textperiodcentered{} 3 & 5/5 \textperiodcentered{} 1.4~s \textperiodcentered{} 3 \\
429, \texttt{Retry-After: 60} & 5/5 \textperiodcentered{} 0.004~s \textperiodcentered{} 1 & 5/5 \textperiodcentered{} 0.002~s \textperiodcentered{} 1 \\
no response for 70~s & 5/5 \textperiodcentered{} 10.3~s \textperiodcentered{} 3 & 5/5 \textperiodcentered{} 10.3~s \textperiodcentered{} 3 \\
200, body not JSON & 5/5 \textperiodcentered{} 0.005~s \textperiodcentered{} 1 & 5/5 \textperiodcentered{} 0.003~s \textperiodcentered{} 1 \\
\bottomrule
\end{tabularx}
\end{table}

Every explanation fell back to the template (Table~\ref{tab:4}). The time to fall back is set by the guard's policy rather than by the framework: three attempts for the transient failures, none for a malformed success, and an immediate stop when the provider asks for a longer wait than the policy allows. A triage loop that waits for its explanation is therefore delayed by at most the policy's budget, and in the worst case measured here by 10.3~s.

\section{Discussion}

\textbf{Why the model should not decide.} On the same evidence, the detectors were right in 90~\% of incidents and the best of four models in 59~\%. The models' errors are not random: they under-recognise data-path corruption and, given readings alone, collapse to one answer. A larger model did not close the gap; the 32~B model attributed no better than the 3~B model from evidence. The detectors carry knowledge the evidence lines only summarise --- the plausible-rate bound, the calibrated noise, the order in which the rule weighs its tests --- and the experiment shows that a model given the summary does not recover it.

\textbf{What the gate is for.} With explicit rules in the prompt, the models rarely broke them, so the gate rejected almost nothing. Its value is that it makes the properties hold by construction rather than by the model's compliance: if a model, a prompt or a framework version changes, an explanation that names another cause, invents a number or recommends a misfitting check still cannot reach the operator. The rules are cheap --- no second model call --- because the task is narrow.

\textbf{What the gate does not check.} R3 verifies that a number occurs in the facts, not that it is used in its role: two explanations in the rated sample quoted the 40~s between trigger and decision as the duration of a stuck value and passed. R5 is lexical: it rejected no reply in this run, but a phrase list can both miss an unusual negation and let a paraphrased claim through. Neither semantic faithfulness nor usefulness to an operator is measured automatically. The author's rating (Section IV.B) found four unfaithful explanations among 50 the rules had accepted, all from one 7~B model and all on instrument decisions, where each named a mechanism the evidence does not show; a rule could check the mechanism as well as the cause, but only for causes whose evidence names one. Usefulness to an operator was not rated.

\textbf{Choice of model.} For the wording task, the trade is between specificity and latency: a 3~B model answers in 0.23~s with generic text, a 32~B model in 15~s with specific text. A deployment can use the small model inline and the large one asynchronously, or present the template at once and replace it when the explanation arrives; the gate and the fallback make both safe.

\textbf{Limits.} The incidents are synthetic (the benchmark of~\cite{r4}); real channels add context --- maintenance logs, neighbouring channels --- that could help a model attribute and that the detectors do not use. Four open models were run locally with greedy decoding; larger hosted models were not evaluated. The prompts were fixed on 65 development incidents; other prompts could change the attribution results, although not the guarantees of the gate. The explanations were rated by one rater, the author, who also built the system; no agreement between raters can be reported, and a rating by operators remains to be done.

\section{Conclusion}

Keeping the decision in deterministic detectors and giving a language model only the wording is supported by measurement: from the detectors' own evidence, four local models attributed telemetry departures correctly in 47--59~\% of 260 incidents, against 90~\% for the detectors, and from raw readings in 5--41~\%. As explainers, the same models passed six mechanical grounding rules in 1,296 of 1,300 cases, never contradicted the decided cause --- in the author's rating of 50 of them, 46 were faithful to the evidence --- and differed mainly in how specific they were --- from 14~\% of explanations quoting a measurement for a 3~B model to 65~\% for a 32~B model. Behind a reliability layer, a failing model server costs the operator nothing but the prose: every explanation fell back to a template within 10.3~s. The explainer is framework-independent; Spring AI and LangChain4j are interchangeable behind its port.

\section*{Data and code availability}

The explainer, the grounding rules, both framework adapters, the experiment harness, every result file and the figure scripts are available at \url{https://github.com/alxkm/triage-explain} under the MIT license; the incidents are generated by the benchmark of~\cite{r4} at commit \texttt{3246e44}, and the reliability layer is that of~\cite{r6}. The results in this paper were produced on 18 October 2026; the repository records the model digests and the runtime environment of each experiment.

\section*{Competing interests}

The author contributes to the open-source projects Spring AI and LangChain4j, through which the models are reached, is the author of the benchmark~\cite{r4} and of the reliability layer~\cite{r6} used here, and contributed the library components~\cite{r5} and the fixes to Eclipse Milo and Apache PLC4X cited in Section II. The work received no external funding.

\FloatBarrier
\makeatletter\ifx\@dbldeferlist\@empty\else\clearpage\fi\makeatother


\begin{thebibliography}{19}

\bibitem{r1} Y. Zheng, C. Dong, J. Wang, A. Topalov, O. Nechai, A. Klymenko, S. Nekrasov, and V. Knysh, ``Information system for operational monitoring of the city's water environment based on IoT technologies,'' \emph{Comptes rendus de l'Acad\'emie bulgare des Sciences}, vol. 78, no. 5, pp. 708--717, 2025, doi: 10.7546/CRABS.2025.05.09.

\bibitem{r2} M. Madinov, A. Topalov, O. Klymenko, V. Dyvak, D. Kropyvnytskyi, and T. Humeniuk, ``Bluemix IIoT system for monitoring production equipment,'' in \emph{Proc. 13th IEEE Int. Conf. Intelligent Data Acquisition and Advanced Computing Systems (IDAACS)}, Gliwice, Poland, 2025, pp. 1335--1340, doi: 10.1109/IDAACS68557.2025.11322007.

\bibitem{r3} V. Khoda, N. Leshchuk, A. Topalov, S. Robotko, O. Klymenko, and S. Nekrasov, ``Computerized lathe control system based on Internet of Things technology,'' in \emph{Proc. 14th Int. Conf. Advanced Computer Information Technologies (ACIT)}, Ceske Budejovice, Czech Republic, 2024, pp. 674--677, doi: 10.1109/ACIT62333.2024.10712548.

\bibitem{r4} O. Klymenko, ``Distinguishing process change, instrument fault and data-path corruption in streaming telemetry: A reference implementation and benchmark,'' arXiv:2610.18904 [eess.SP], Oct. 2026, doi: 10.48550/arXiv.2610.18904. Code: \url{https://github.com/alxkm/telemetry-triage}

\bibitem{r5} TheAlgorithms/Java: streaming detectors contributed by the author, among them TheAlgorithms/Java\#7592 (Kalman filter), \#7593 (CUSUM), \#7598 (Hampel filter), \#7600 (EWMA chart), \#7606 (ADWIN) and \#7612 (generalized ESD), Sept. 2026. [Online]. Available: \url{https://github.com/TheAlgorithms/Java}

\bibitem{r6} O. Klymenko, ``Consistent LLM call handling across Java frameworks: A reference implementation and benchmark,'' arXiv:2610.25480 [cs.LG], Oct. 2026, doi: 10.48550/arXiv.2610.25480. Code: \url{https://github.com/alxkm/llm-call-guard}

\bibitem{r7} O. Klymenko, \texttt{triage-explain}: source code and results, 2026. [Online]. Available: \url{https://github.com/alxkm/triage-explain}

\bibitem{r8} T. Ahmed, S. Ghosh, C. Bansal, T. Zimmermann, X. Zhang, and S. Rajmohan, ``Recommending root-cause and mitigation steps for cloud incidents using large language models,'' in \emph{Proc. 45th IEEE/ACM Int. Conf. Software Engineering (ICSE)}, 2023, pp. 1737--1749, doi: 10.1109/ICSE48619.2023.00149.

\bibitem{r9} Y. Chen, H. Xie, M. Ma, Y. Kang, X. Gao, L. Shi, Y. Cao, X. Gao, H. Fan, M. Wen, J. Zeng, S. Ghosh, X. Zhang, C. Zhang, Q. Lin, S. Rajmohan, D. Zhang, and T. Xu, ``Automatic root cause analysis via large language models for cloud incidents,'' in \emph{Proc. 19th European Conf. Computer Systems (EuroSys)}, 2024, pp. 674--688, doi: 10.1145/3627703.3629553.

\bibitem{r10} S. Alnegheimish, L. Nguyen, L. Berti-Equille, and K. Veeramachaneni, ``Can large language models be anomaly detectors for time series?'' in \emph{Proc. IEEE 11th Int. Conf. Data Science and Advanced Analytics (DSAA)}, 2024, pp. 1--10, doi: 10.1109/DSAA61799.2024.10722786.

\bibitem{r11} N. Gruver, M. Finzi, S. Qiu, and A. G. Wilson, ``Large language models are zero-shot time series forecasters,'' in \emph{Advances in Neural Information Processing Systems 36}, 2023, pp. 19622--19635, doi: 10.52202/075280-0861.

\bibitem{r12} Z. Ji, N. Lee, R. Frieske, T. Yu, D. Su, Y. Xu, E. Ishii, Y. J. Bang, A. Madotto, and P. Fung, ``Survey of hallucination in natural language generation,'' \emph{ACM Computing Surveys}, vol. 55, no. 12, pp. 1--38, 2023, doi: 10.1145/3571730.

\bibitem{r13} P. Manakul, A. Liusie, and M. Gales, ``SelfCheckGPT: Zero-resource black-box hallucination detection for generative large language models,'' in \emph{Proc. Conf. Empirical Methods in Natural Language Processing (EMNLP)}, 2023, pp. 9004--9017, doi: 10.18653/v1/2023.emnlp-main.557.

\bibitem{r14} S. Min, K. Krishna, X. Lyu, M. Lewis, W.-t. Yih, P. W. Koh, M. Iyyer, L. Zettlemoyer, and H. Hajishirzi, ``FActScore: Fine-grained atomic evaluation of factual precision in long form text generation,'' in \emph{Proc. Conf. Empirical Methods in Natural Language Processing (EMNLP)}, 2023, pp. 12076--12100, doi: 10.18653/v1/2023.emnlp-main.741.

\bibitem{r15} E. Reiter and R. Dale, \emph{Building Natural Language Generation Systems}. Cambridge: Cambridge University Press, 2000, doi: 10.1017/CBO9780511519857.

\bibitem{r16} A. Gatt and E. Krahmer, ``Survey of the state of the art in natural language generation: Core tasks, applications and evaluation,'' \emph{Journal of Artificial Intelligence Research}, vol. 61, pp. 65--170, 2018, doi: 10.1613/jair.5477.

\bibitem{r17} Eclipse Milo, an open-source implementation of OPC UA (IEC 62541), version 1.1.7. [Online]. Available: \url{https://github.com/eclipse-milo/milo}

\bibitem{r18} Apache PLC4X, the industrial IoT adapter. [Online]. Available: \url{https://github.com/apache/plc4x}

\bibitem{r19} Ollama, version 0.40.5. [Online]. Available: \url{https://github.com/ollama/ollama}

\end{thebibliography}
\end{document}